%% file: 0_main.tex
\documentclass[lettersize,journal]{IEEEtran}
\usepackage{amsmath,amsfonts}
\usepackage{amsthm}
\usepackage{algorithm}
\usepackage{algorithmic}
\usepackage{array}
\newcolumntype{P}[1]{>{\centering\arraybackslash}p{#1}}
\newcolumntype{M}[1]{>{\centering\arraybackslash}m{#1}}
\usepackage[caption=false,font=normalsize,labelfont=sf,textfont=sf]{subfig}
\usepackage{textcomp}
\usepackage{stfloats}
\usepackage{url}
\usepackage{verbatim}
\usepackage{graphicx}
\def\BibTeX{{\rm B\kern-.05em{\sc i\kern-.025em b}\kern-.08em
    T\kern-.1667em\lower.7ex\hbox{E}\kern-.125emX}}
\usepackage{balance}
\usepackage{booktabs}
\usepackage{multirow}
\usepackage{caption}
\usepackage{xcolor}
\usepackage{authblk}
\begin{document}
\title{\hspace{-0.36cm}Is Complexity Required for Neural Network Pruning? A Case Study on Global Magnitude Pruning}

\author{Manas~Gupta,
        Efe~Camci, 
        Vishandi~Rudy~Keneta,
        Abhishek~Vaidyanathan,
        Ritwik~Kanodia,
        Chuan-Sheng~Foo,
        Min~Wu and
        Jie~Lin
\IEEEcompsocitemizethanks{
		\IEEEcompsocthanksitem Manas Gupta, Efe Camci, Chuan-Sheng Foo, Min Wu and Jie Lin\textsuperscript{1} are with the Institute for Infocomm Research (I$^2$R), A*STAR, 138632, Singapore. Email: \{{\protect\url{manas_gupta}, \protect\url{efe_camci},  \protect\url{foo_chuan_sheng}, \protect\url{wumin}\}@i2r.a-star.edu.sg}. \textsuperscript{1}Work done while at I2R \protect\url{{jie.dellinger@gmail.com}}.
		\IEEEcompsocthanksitem Vishandi Rudy Keneta is with the School of Computing (SoC), National University of Singapore (NUS), 119077, Singapore. Email: \protect\url{e0407662@u.nus.edu}.
		\IEEEcompsocthanksitem Abhishek Vaidyanathan and Ritwik Kanodia\textsuperscript{2} are with the School of Computer Science and Engineering (SCSE), Nanyang Technological University (NTU), 639798, Singapore. Email: \protect\url{abhishek033@e.ntu.edu.sg}. \textsuperscript{2}Work done while at NTU \protect\url{{ritwikkanodiain@gmail.com}}.
}
}

\maketitle
\begin{abstract}

Pruning neural networks has become popular in the last decade when it was shown that a large number of weights can be safely removed from modern neural networks without compromising accuracy. Numerous pruning methods have been proposed since, each claiming to be better than prior art, however, at the cost of increasingly complex pruning methodologies. These methodologies include utilizing importance scores, getting feedback through back-propagation or having heuristics-based pruning rules amongst others. In this work, we question whether this pattern of introducing complexity is really necessary to achieve better pruning results. We benchmark these SOTA techniques against a simple pruning baseline, namely, Global Magnitude Pruning (Global MP), that ranks weights in order of their magnitudes and prunes the smallest ones. Surprisingly, we find that vanilla Global MP performs very well against the SOTA techniques. When considering sparsity-accuracy trade-off, Global MP performs better than all SOTA techniques at all sparsity ratios. When considering FLOPs-accuracy trade-off, some SOTA techniques outperform Global MP at lower sparsity ratios, however, Global MP starts performing well at high sparsity ratios and performs very well at extremely high sparsity ratios. Moreover, we find that a common issue that many pruning algorithms run into at high sparsity rates, namely, layer-collapse, can be easily fixed in Global MP. We explore why layer collapse occurs in networks and how it can be mitigated in Global MP by utilizing a technique called Minimum Threshold. We showcase the above findings on various models (WRN-28-8, ResNet-32, ResNet-50, MobileNet-V1 and FastGRNN) and multiple datasets (CIFAR-10, ImageNet and HAR-2). Code is available at https://github.com/manasgupta-1/GlobalMP. 

\end{abstract}


\input{1_Introduction}
\input{2_RelatedWork}
\input{3_Approach}
\input{4_Results}
\input{5_Discussion}
\input{6_Conclusion}
\bibliographystyle{IEEEtran}
\bibliography{IEEEabrv, references.bib}

\input{7_Appendix}

\end{document}

%% file: 1_Introduction.tex
\section{Introduction}
\label{introduction}

Scaling-up the size of neural networks is becoming a popular way to increase model performance {\cite{bert}}, {\cite{gale2019state}}, {\cite{scalinglaws}}. However, this poses a significant cost to the environment {\cite{strubell2019energy}} and makes deployment on edge devices difficult {\cite{bonatti2020autonomous}. Neural network pruning has thus emerged as an essential tool to reduce the size of modern-day neural networks. New methods have utilized a myriad of pruning techniques consisting of gradient-based methods, sensitivity to or feedback from an objective function, distance or similarity measures, regularization-based techniques, amongst others. The state-of-the-art (SOTA) pruning techniques use complex rules like iterative pruning and re-growth of weight parameters using heuristics rules every few hundred iterations, as in DSR \cite{ICML-2019-MostafaW}. SM \cite{sparse_momentum} uses sparse momentum that benefits from exponentially smoothed gradients (momentum) to find layers and weights that reduce error and then redistribute the pruned weights across layers using the mean momentum magnitude of each layer. For each layer, sparse momentum grows the weights using the momentum magnitude of zero-valued weights. Another popular SOTA technique, RigL \cite{pmlr-v119-evci20a}, also iteratively prunes and re-grows weights every few iterations. They use uniform or Erdos-Renyi-Kernel (ERK) for pruning connections and re-grow connections based on the highest magnitude gradients. Among recent techniques, DPF \cite{Lin2020Dynamic} uses dynamic allocation of the sparsity pattern and incorporates a feedback signal to re-activate prematurely pruned weights, while STR \cite{pmlr-v119-kusupati20a} utilises Soft Threshold Reparameterization and uses back-propagation to find sparsity ratios for each layer. 

Despite the high number of new pruning algorithms proposed, the tangible benefits of many of them are still questionable. For instance, recently it has been shown that many pruning at initialization (PAI) schemes do not perform as well as expected \cite{frankle2021pruning}. In that paper, it is shown through a number of experiments that these PAI schemes are actually no better than random pruning, which is one of the most naive pruning baselines with no complexity involved. This finding indeed raises another question in our minds: if a well designed PAI does not even match the performance of random pruning, can simple pruning approaches like global pruning or their variants outperform other existing algorithms? In this work, we question the trend of proposing increasingly complex pruning algorithms and evaluate whether such complexity is really required to achieve superior results. We benchmark popular state-of-the-art (SOTA) pruning techniques against a naive pruning baseline, namely, Global Magnitude Pruning (Global MP). Global MP ranks all the weights in a neural network by their magnitudes and prunes off the smallest ones (Fig.~\ref{fig:workflow}). 

Despite its simplicity, Global MP has not been comprehensively analyzed and evaluated in the literature. Although, some prior works have used Global MP as a baseline \cite{frankle2018lottery, NEURIPS2019_a4613e8d, blalock2020state, NEURIPS2020_46a4378f, Renda2020Comparing, lee2021layeradaptive}, they missed out on conducting rigorous experiments with it; for example, in settings of both gradual and one-shot pruning or comparing it with SOTA. Similarly, many SOTA papers do not use Global MP for benchmarking and miss out on capturing its remarkable performance \cite{pmlr-v119-evci20a, pmlr-v119-kusupati20a, Zhu2018ToPO, gale2019state, DNW}. We bridge this gap in evaluating the efficacy of Global MP and demonstrate its superior performance under multiple experimental conditions. 

We show that with regards to the trade-off between sparsity and accuracy, Global MP consistently outperforms all state-of-the-art (SOTA) techniques across various sparsity ratios. In terms of the trade-off between FLOPs and accuracy, certain SOTA techniques exhibit superior performance than Global MP at lower sparsity ratios. However, Global MP demonstrates significant efficacy at higher sparsity ratios and excels particularly well at extremely high sparsity levels. While achieving such performance, Global MP does not require any additional algorithm-specific hyper-parameters to be tuned. We also shed light into a potential problem with pruning, known as layer-collapse, whereby an entire layer is pruned away, leading to a drastic loss in accuracy. The fix for it in Global MP is simple through introducing a Minimum Threshold (MT) to retain a minimum number of weights in every layer. We conduct experiments on WRN-28-8, ResNet-32, ResNet-50, MobileNet-V1, and FastGRNN models, and on CIFAR-10, ImageNet, and HAR-2 datasets. We test Global MP for both unstructured and structured as well as one-shot and gradual settings, and share our findings.

\begin{figure*}[!h]
    \centering
    \includegraphics[trim=0cm 5.5cm 0cm 8cm,clip,width=0.75\textwidth]{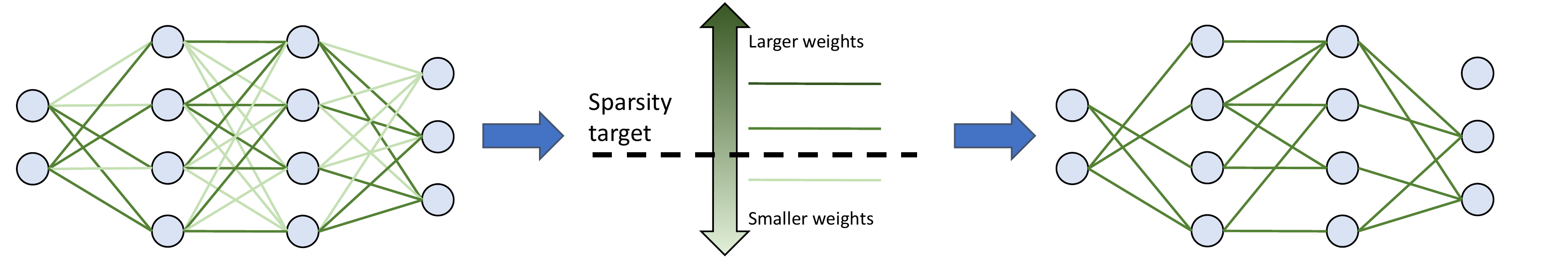}
    \caption{Illustration of how Global MP works. Global MP ranks all the weights in a network by their magnitudes and prunes off the smallest weights until the target sparsity is met. Light green weights refer to the smaller-magnitude weights which are pruned off. A pruned network consisting of larger-magnitude weights (dark green weights) is obtained after the process.}
    \label{fig:workflow}
\end{figure*}

%% file: 2_RelatedWork.tex
\section{Related Work}
\label{related_work}

Compression of neural networks has become an important research area due to the rapid increase in size of neural networks \cite{brown2020language}, the need for fast inference \cite{camci2020deep}, application to real-world tasks \cite{9516010, 8818358, 8756206, Liu2021ARA, 8693518} and concerns about the carbon footprint of training large neural networks \cite{strubell2019energy}. Over the years, several compression techniques have emerged in the literature \cite{cheng2017survey, 9478787}, such as quantisation, factorisation, attention, knowledge distillation, architecture search and pruning \cite{almahairi2016dynamic,ashok2017n2n,i2016squeezenet,pham2018efficient}. As compared to other categories, pruning is more general in nature and has shown strong performance \cite{gale2019state}.

Many pruning techniques have been developed over the years, which use first or second order derivatives \cite{NIPS1989_250,NIPS1992_647, bert}, gradient based methods \cite{lee2018snip, Wang2020Picking, acdc}, sensitivity to or feedback from some objective function \cite{Lin2020Dynamic, molchanov2016pruning, LIU2020Dynamic, jorge2021progressive, 9097925}, distance or similarity measures \cite{Srinivas_2015}, Bayesian optimisation {\cite{automatedfilter}}, regularization-based techniques \cite{autobalanced, pmlr-v119-kusupati20a, ContinuousSparsification2020, wang2021neural, 9398648}, and magnitude-based criterion \cite{pmlr-v119-evci20a, lee2021layeradaptive, Zhu2018ToPO, Strom97sparseconnection, Park2020LookaheadAF}. A key trick has been discovered in \cite{han2015learning} to iteratively prune and retrain a network, thereby preserving high accuracy. Runtime Neural Pruning \cite{NIPS2017_6813} attempts to use reinforcement learning (RL) for compression by training an RL agent to select smaller sub-networks during inference. \cite{he2018amc} design the first approach using RL for pruning. However, RL training approaches typically require additional RL training budgets and careful RL action and state space design \cite{gupta2020learning, qlp}.

Global MP on the other hand works by ranking all the parameters in a network by their absolute magnitudes and then pruning the smallest ones. It is therefore, quite intuitive, logical and straightforward. It is also not to be confused with methods that utilize global pruning but do not conduct magnitude pruning, for example, SNIP {\cite{lee2018snip}}. These methods use complex criteria, first to determine the saliency of the weights globally and then apply pruning. We present here an in-depth comparison of the SOTA techniques vs. Global MP. Gradual Magnitude Pruning (GMP) {\cite{Zhu2018ToPO}} uses a uniform pruning schedule and prunes each layer by the same amount, thereby, not taking into account the relative importance of layers. Global MP on the other hand prunes every layer differently. Dynamic Sparse Reparameterization (DSR) {\cite{ICML-2019-MostafaW}} prunes and regrows weights every few hundred iterations. It also uses Global MP to prune the weights. However, it imposes some additional heuristic-based constrains on the pruning process, such as, not pruning some selected layers in the network. In case the weights to regrow outnumber the capacity of a layer, it then uses additional heuristics to redistribute the additional weights. These kinds of heuristics add both complexity and limit the potential pruning actions that can be taken. Discovering Neural Wirings (DNW) {\cite{DNW}} focuses on learning connectivity of channels in a network. It is not primarily a pruning technique and is more akin to Neural Architecture Search (NAS).

Sparse Momentum (SM) {\cite{sparse_momentum}} uses a heuristic-based approach to prune and regrow weights. They use average momentum to assess importance of every layer and assign parameters accordingly. Similar to DSR, a certain number of layers are never pruned and in cases of regrowth exceeding capacity of a layer, additional heuristics are used to redistribute weights. This method is also computationally intensive as gradients need to be stored in memory and additional FLOPs are required to calculate mean momentum per parameter. Hence, it is not as efficient as Global MP, which is computationally less expensive and does not rely on heuristics for pruning. Rigging the Lottery (RigL) {\cite{pmlr-v119-evci20a}} allocates sparsity based on the number of parameters in a layer. Hence, importance of a layer is based on its size, which is not necessarily an accurate measure for all cases. Dynamic Pruning with Feedback (DPF) {\cite{Lin2020Dynamic}} also uses Global MP for pruning. However, it imposes an additional constraint of keeping the last layer fully dense. It is thus, not as flexible as pure Global MP that allows all layers to be pruned.

Soft Threshold Reparameterization (STR) {\cite{pmlr-v119-kusupati20a}} is a regularization-based technique that subtracts a certain value from the weights in each epoch. The exact sparsity target cannot be controlled in STR, and it requires heavy tuning of the hyper-parameters to reach a required sparsity target. Hence, it is not as flexible and generalizable as Global MP. Matrix-Free Approximations of Second-Order Information (M-FAC) {\cite{mfac}} and its pre-cursor (WoodFisher) {\cite{woodfisher}} use an approximation of the second order Hessian to prune weights. Hessian-based pruning is principled but computationally intractable. Therefore, approximations need to be made for the Hessian. These approximations result in not-so-accurate importance scores for the weights, while such approximations are not needed in Global MP. Thus, Global MP is generally more principled, heuristic-free, and computationally inexpensive compared to SOTA methods and might be the underlying reason for its superior performance.

%% file: 3_Approach.tex
\section{Method}
\label{approach}

In this section, we explain how Global MP works by describing its key components. We also introduce a simple thresholding mechanism, called \textit{Minimum Threshold (MT)}, to avoid the issue of layer-collapse at high sparsity levels. 

\subsection{Global Magnitude Pruning (Global MP)}
\label{algo:gp}

Global MP is a magnitude-based pruning approach, whereby weights larger than a certain threshold are kept, and weights smaller than the threshold are pruned across a neural network. The threshold is calculated based on the target sparsity rate and is not a hyper-parameter that needs to be tuned or learnt. Given a target sparsity rate $\kappa_{target}$, the threshold $t$ is simply calculated as the weight magnitude that serves as a separation point between the smallest $\kappa_{target}$ percent of weights and the rest, once all weights are sorted into an array based on their magnitude. Formally, for a calculated threshold $t$ and each individual weight $w$ in any layer, the new weight $w_{new}$ is defined as follows:
\begin{equation}
    w_{new} = \begin{cases} 
      0 & |w| < t, \\
      w & otherwise. \\
   \end{cases}
\end{equation}

In Global MP, a single threshold is set for the entire network based on the target sparsity for the network. This is in contrast to layer-wise pruning, in which different threshold values have to be searched for each layer individually. In the case of uniform pruning on the other hand, a threshold for each layer needs to be calculated based on the sparsity target assigned to the layers uniformly across the network. In this aspect, Global MP is more efficient than layer-wise or uniform pruning because the threshold does not need to be searched or calculated for every layer individually.

\subsection{Minimum Threshold (MT)}
\label{algo:MT}

The Minimum Threshold (MT) refers to the fixed number of weights that are preserved in every layer of the neural network post pruning. The MT is a scalar value that is fixed before the start of the pruning cycle. The weights in a layer are sorted by their magnitude and the largest MT number of weights are preserved. For instance, an MT of 500 implies that 500 of the largest weights in every layer need to be preserved post pruning. If a layer originally has a smaller number of weights than the MT number, then all the weights of that layer will be preserved. This corresponds to:

\begin{equation}
{\|W_l\|_0}
 \geq \begin{cases} 
      \sigma & \text{if } m \geq \sigma_l, \\
      m & \text{otherwise.} \\
    \end{cases}
\label{eq:MT}
\end{equation}

The term $W_l \in \mathbb{R}^{m}$ denotes the weight vector for layer $l$, $\sigma$ is the MT value in terms of the number of weights and ${\|W_l\|_0}$ indicates the number of non-zero elements in $W_l$. We explain in the next subsection how the actual pruning using MT is implemented.

\subsection{The Pruning Workflow}

The pruning pipeline for Global MP consists of pruning a model until the desired sparsity target is met and training or fine-tuning it a the specified number of epochs. It supports both one-shot and gradual pruning settings as well as with or without MT. The users may choose any pruning setting as per their use case. The procedure starts by selecting a pre-trained model in one-shot pruning, or an untrained model in gradual pruning. Next, the sparsity of the model is checked and if the sparsity is lower than the target sparsity, then the model is pruned using either vanilla Global MP or Global MP with MT, as per the choice of the user. Once, the model is pruned, it is trained for the case of gradual pruning or fine-tuned for the case of one-shot pruning. The Global MP framework allows the flexibility for previously pruned weights to regrow, if they become more active in the later epochs, for the case of gradual pruning. No hard pruning is done whereby weights are permanently zeroed out. The pruning mask is calculated afresh in each epoch thereby allowing previously pruned weights to regrow. The above procedure repeats until the final epoch is reached. For the case of one-shot pruning, the later epochs are just used for doing fine-tuning as the pruning happens in one-go in the first epoch itself. This finishes the procedure and the final result is a pruned and trained (or fine-tuned) model. See Appendix for pseudocode.

%% file: 4_Results.tex
\section{Experiments}
\label{Results}
Below we describe experiments related to Global MP compared to state-of-the-art (SOTA) pruning algorithms.

\subsection{Comparison with SOTA}

We compare Global MP with various popular SOTA algorithms that are well known for pruning, such as SNIP \cite{lee2018snip}, SM \cite{sparse_momentum}, DSR \cite{ICML-2019-MostafaW}, DPF \cite{Lin2020Dynamic}, GMP \cite{Zhu2018ToPO}, DNW \cite{DNW}, RigL \cite{pmlr-v119-evci20a}, and STR \cite{pmlr-v119-kusupati20a}. These include a broad spectrum of methods involving iteratively pruning and re-growing weights every few iterations, pruning at initialization, using gradients and feedback signals for pruning, and pruning using regularization. We report results from these algorithms whenever they report results for the specific dataset that is being experimented upon. See Appendix for hyper-parameters.

\subsubsection{CIFAR-10}
\label{sota_cifar10}
We conduct experiments to compare Global MP to SOTA pruning algorithms on the CIFAR-10 dataset. We compare One-shot Global MP with four algorithms in this case: SNIP \cite{lee2018snip}, SM \cite{sparse_momentum}, DSR \cite{ICML-2019-MostafaW}, and DPF \cite{Lin2020Dynamic}. We report results on two popular and widely pruned network architectures, namely, WideResNet-28-8 (WRN-28-8) and ResNet-32 \cite{DBLP:journals/corr/HeZRS15}. For both architectures, we start off with the original model having the same initial accuracy as the other algorithms to have a fair comparison. For WRN-28-8 experiments Table~\ref{table:cifar10}, Global MP performs better than the rest of the competitors at both 90\% and 95\% sparsity levels. Global MP outperforms DSR and SM in all cases, because of their additional heuristics-based constraints which limit the selection of layers to be pruned. As for ResNet-32 (Table~\ref{table:cifar10}), Global MP outperforms at 95\% sparsity and is the second best at 90\% sparsity. Global MP outperforms DSR and SM for all sparsity levels in this case as well. This is an indication of the capabilities of Global MP as compared to the other algorithms, while featuring no added complexity.
\begin{table}[t!]
\small
\centering
\begin{tabular}{p{1.5cm}p{1.1cm}p{2.1cm}p{2.1cm}}
\toprule
\multirow{1}{*}{Method} & Sparsity & WRN-28-8 Acc. & ResNet-32 Acc.\\
\midrule
Baseline & 0.0\% & 96.06\% & 93.83 $\pm$ 0.12 \%\\
\midrule
SNIP & 90\% & $95.49 \pm 0.21\%$ & 90.40 $\pm$ 0.26\%\\
SM & 90\% & $95.67 \pm 0.14\%$ & 91.54 $\pm$ 0.18\%  \\
DSR & 90\% & $95.81 \pm 0.10\%$ & 91.41 $\pm$ 0.23\%  \\
DPF & 90\% & $96.08 \pm 0.15\%$ & 92.42 $\pm$ 0.18\%\\
\textbf{Global MP} & 90\% & $\textbf{96.30} \pm \textbf{0.03\%}$ & $\textbf{92.67} \pm \textbf{0.03\%}$\\
\midrule
SNIP & 95\% & $94.93 \pm 0.13\%$ & 87.23 $\pm$ 0.29\%  \\
SM & 95\% & $95.64 \pm 0.07\%$ & 88.68 $\pm$ 0.22\% \\
DSR & 95\% & $95.55 \pm 0.12\%$ & 84.12 $\pm$ 0.32\% \\
\textbf{DPF} & 95\% & $95.98 \pm 0.10\%$ & \textbf{90.94 $\pm$ 0.35\%} \\
\textbf{Global MP} & 95\% & $\textbf{96.16} \pm \textbf{0.02\%}$ & 90.65 $\pm$ 0.13\%\\
\bottomrule
\end{tabular}
\captionof{table}{Results of SOTA pruning algorithms on WideResNet-28-8 and ResNet-32 on CIFAR-10. The bold font denotes algorithm with the best performance. Global MP outperforms or yields comparable performance to other algorithms.}
\label{table:cifar10}
\end{table}

\subsubsection{ImageNet}
\label{sota_imagenet}
Following the favorable performance on CIFAR-10 dataset, we benchmark Global MP against other competitors in the literature over ImageNet dataset. This is a highly challenging dataset as compared to CIFAR-10, featuring around 1.3 million RGB images with 1,000 classes. Using this dataset, we compare Global MP with SOTA algorithms like GMP \cite{Zhu2018ToPO}, DSR \cite{ICML-2019-MostafaW}, DNW \cite{DNW}, SM  \cite{sparse_momentum}, RigL \cite{pmlr-v119-evci20a}, WoodFisher {\cite{woodfisher}}, MFAC {\cite{mfac}}, DPF \cite{Lin2020Dynamic}, and STR \cite{pmlr-v119-kusupati20a}. The two network architectures we use for this comparison are ResNet-50 and MobileNet-V1 \cite{howard2017mobilenets}, the two most popular architectures for benchmarking pruning algorithms on ImageNet \cite{blalock2020state}. For ResNet-50, we add in an additional experimental setting of gradual Global MP to provide more thorough comparison with SOTA methods. We again start from the same initial accuracy for the non-pruned models for all algorithms, either by matching the results in their original papers or reproducing their results whenever their code is available. We sample four sparsity levels ranging from low sparsity (80\%) to extreme sparsity (98\%) to provide a comprehensive snapshot across different sparsity levels.

The remarkable performance of Global MP becomes clearly visible in ResNet-50 over ImageNet experiments. As can be seen from Table~\ref{table:resnet50_imagenet}, for the sparsity-accuracy trade-off, Global MP outperforms or achieves comparable accuracy as all the other competitors in every sparsity level from 80\% to 98\% (see results in bold). MFAC performs closely for 95\% sparsity however, its not a like-for-like comparison as their sparsity is slightly lower (95\%) compared to Global MP (95.3\%). We take the upper bound sparsity target for each sparsity level for Global MP to match the method with the highest reported sparsity in that sparsity level. For the case of extreme sparsity (98\%), Global MP surpasses the second best algorithm (STR) by a large margin of 5.11\%. For the FLOPs-accuracy trade-off, the comparison is more difficult as the methods report different FLOPs targets and a like-for-like comparison cannot be done simply. However, an experienced practitioner can roughly gauge the efficacy of the methods based on the ratio between the additional FLOPs pruned to decrease in accuracy of the methods. Based on this we find that certain SOTA techniques exhibit superior FLOPs-accuracy performance than Global MP at lower sparsity ratios of 80\% and 90\%. However, Global MP becomes competitive at 95\% sparsity and performs very well at the extreme sparsity rate of 98\% sparsity, gaining 5.11\% accuracy vs. a drop of 2\% FLOPs vis-a-vis the second best method (STR).

We also find that gradual Global MP generally performs better than one-shot Global MP at high and extremely high sparsity ratios. This is because the pruning mask is allowed to change multiple times in gradual Global MP compared to only once in one-shot Global MP, and hence, converges to a more optimized value. Most SOTA methods also follow the same approach whereby they allow the pruning mask to change each epoch or sometimes multiple times in an epoch. Overall, Global MP outperforms all SOTA algorithms on sparsity-accuracy trade-off and comes in second, after STR, for FLOPs-accuracy trade-off. It is an important finding that such a simple algorithm like Global MP can outperform other SOTA competitors that incorporate very complex design choices or computationally demanding procedures.

\begin{table}[t!]
\small
\centering
\begin{tabular}[t]{p{3cm}p{0.9cm}p{0.8cm}p{1cm}p{0.8cm}}
\toprule
\multirow{1}{*}{Method} & Top-1 Acc & Params & Sparsity & FLOPs pruned\\
\midrule
ResNet-50 & 77.0\% & 25.6M\ & 0.00\% & 0.0\%\\
\midrule
GMP & 75.60\% & 5.12M\ & 80.00\% & 80.0\%\\
DSR*\#\ & 71.60\% & 5.12M\ & 80.00\% & 69.9\%\\
DNW & 76.00\% & 5.12M\ & 80.00\% & 80.0\%\\
SM & 74.90\% & 5.12M\ & 80.00\% & -\\
SM + ERK & 75.20\% & 5.12M\ & 80.00\% & 58.9\%\\
RigL* & 74.60\% & 5.12M\ & 80.00\% & 77.5\%\\
RigL + ERK & 75.10\% & 5.12M\ & 80.00\% & 58.9\%\\
DPF & 75.13\% & 5.12M\ & 80.00\% & 80.0\%\\
\underline{STR} & \underline{76.19\%} & 5.22M\ & 79.55\% & \underline{81.3\%}\\
\textbf{Global MP (One-shot)} & \textbf{76.84\%} & 5.12M & \textbf{80.00\%} & 72.4\%\\
Global MP (Gradual) & 76.12\% & 5.12M & 80.00\% & 76.7\%\\
\midrule
GMP & 73.91\% & 2.56M\ & 90.00\% & 90.0\%\\
DNW & 74.00\% & 2.56M\ & 90.00\% & 90.0\%\\
SM & 72.90\% & 2.56M\ & 90.00\% & 60.1\%\\
SM + ERK & 72.90\% & 2.56M\ & 90.00\% & 76.5\%\\
RigL* & 72.00\% & 2.56M\ & 90.00\% & 87.4\%\\
RigL + ERK & 73.00\% & 2.56M\ & 90.00\% & 76.5\%\\
DPF\# & 74.55\% & 4.45M\ & 82.60\% & 90.0\%\\
\underline{STR} & \underline{74.73\%} & 3.14M\ & 87.70\% & \underline{90.2\%}\\
\textbf{Global MP (One-shot)} & \textbf{75.28\%} & 2.56M & \textbf{90.00\%}  & 82.8\%\\
Global MP (Gradual) & 74.83\% & 2.56M & 90.00\%  & 87.8\%\\
\midrule
GMP & 70.59\% & 1.28M\ & 95.00\% & 95.0\%\\
DNW & 68.30\% & 1.28M\ & 95.00\% & 95.0\%\\
RigL* & 67.50\% & 1.28M\ & 95.00\% & 92.2\%\\
RigL + ERK & 70.00\% & 1.28M\ & 95.00\% & 85.3\%\\
WoodFisher & 72.12\% & 1.28M & 95.00\% & -\\
\textbf{MFAC} & \textbf{72.32\%} & 1.28M & \textbf{95.00\%} & -\\
\underline{STR} & \underline{70.40\%} & 1.27M\ & 95.03\% & \underline{96.1\%}\\
Global MP (One-shot) & 71.56\% & 1.20M & 95.30\% & 89.3\%\\
\underline{\textbf{Global MP (Gradual)}} & \underline{\textbf{72.14\%}} & 1.20M & \textbf{95.30\%} & \underline{93.1\%}\\
\midrule
GMP & 57.90\% & 0.51M\ & 98.00\% & 98.0\%\\
DNW & 58.20\% & 0.51M\ & 98.00\% & 98.0\%\\
STR & 61.46\% & 0.50M & 98.05\% & 98.2\%\\
Global MP (One-shot) & 61.80\% & 0.50M & 98.05\% & 93.7\%\\
\underline{\textbf{Global MP (Gradual)}} & \underline{\textbf{66.57}}\% & 0.50M & \textbf{98.05\%} & \underline{96.2\%}\\
\bottomrule
\end{tabular}
\vspace{3pt}
\captionof{table}{Results on ResNet-50 on ImageNet. Global MP outperforms SOTA pruning algorithms at all sparsity levels for the sparsity-accuracy trade-off and at high sparsity levels for the FLOPs-accuracy trade-off. Bold font denotes best performance for the sparsity-accuracy trade-off while underlined font denotes best performance for FLOPs-accuracy trade-off. * and \# imply the first and the last layer are dense, respectively.}
\vspace*{-0.3cm}
\label{table:resnet50_imagenet}
\end{table}

\begin{table}[t!]
\centering
\small
\begin{tabular}{p{3cm}p{0.9cm}p{0.8cm}p{1cm}p{0.8cm}}
\toprule
\multirow{1}{*}{Method} & Top-1 Acc & Params. & Sparsity & FLOPs pruned\\
\midrule
MobileNet-V1 & 71.95\% & 4.21M\ & 0.00\% & 0.0\%\\
\midrule
GMP & 67.70\% & 1.09M\ & 74.11\% & 71.4\%\\
\underline{STR} & \underline{68.35\%} & 1.04M\ & 75.28\% & \underline{82.2\%}\\
\textbf{Global MP} & \textbf{70.74\%} & 1.04M & \textbf{75.28\%} & 68.9\%\\ 
\midrule
GMP & 61.80\% & 0.46M\ & 89.03\% & 85.6\%\\
\underline{STR} & \underline{61.51\%} & 0.44M\ & 89.62\% & \underline{93.0\%}\\
Global MP & 59.49\% &0.42M\ & 90.00\% & 83.7\%\\
\textbf{Global MP with MT} & \textbf{63.94\%} & 0.42M & \textbf{90.00\%} & 72.9\%\\
\bottomrule
\end{tabular}
\captionof{table}{Results of pruning algorithms on MobileNet-V1 on ImageNet. The bold font denotes the algorithm with the best sparsity-accuracy performance while underlining denotes the best FLOPs-accuracy performance. Global MP with MT surpasses SOTA algorithms on sparsity-accuracy performance.}
\label{table:mobilenetv1_imagenet}
\end{table}

We also test another architecture on ImageNet, MobileNet-V1, which is a much smaller and more efficient architecture than ResNet-50. In this case, strong competitors are limited in the literature; only two of the aforementioned algorithms are able to present competitive results due to the fact that this architecture has less redundancy. We benchmark Global MP with two other competitors at two target sparsity levels: 75\% and 90\%. As can be seen in Table~\ref{table:mobilenetv1_imagenet}, Global MP outperforms SOTA algorithms on the sparsity-accuracy trade-off by a margin of more than 2\% at 75\% sparsity, which is a significant result given how compact MobileNet-V1 is. At 90\% sparsity on the other hand, the same compactness causes Global MP to over-prune certain layers in the network, which result in a significant accuracy drop. This is the above-mentioned problem of layer-collapse, and it is easily rectified when MT is introduced to Global MP. We use an MT value of 0.2\% which is determined using the same grid-search procedure as any other hyper-parameter. Usually values between 0.01\% to 0.3\% work well for MT regardless of models and datasets. See Appendix for an ablation study on MT. We find that MT behaves like a typical hyper-parameter. On increasing the MT value initially, the accuracy increases, until it reaches a maximum value. Thereafter, increasing the MT value leads to a decrease in accuracy. Thus, a suitable value can be found by doing a search over MT values. The accuracy of Global MP at 90\% sparsity goes beyond SOTA again with such a simple fix, and the accuracy margin to the next competitor gets higher than 2\%. For the FLOPs-accuracy trade-off, a like-for-like comparison is again hard to make, but STR seems to perform better, owing to the smaller sparsities that MobileNet is pruned at, as compared to ResNet-50. MT also comes at the cost of a less FLOPs reduction, but it is useful especially for accuracy-critical applications where decreasing the size of the network is still important. All these findings clearly indicate that Global MP is a simple yet competitive pruning algorithm. It delivers top performance on sparsity-accuracy trade-off, and ranks second for FLOPs-accuracy trade-off, despite not having any complex design choices or additional hyper-parameters.

\subsection{Structured pruning and generalizing to other domains and RNN architectures}
\label{rnn}
We experiment with Global MP on other domains and non-convolutional networks as well to measure the generalizability of the algorithm on different domains and network types. We experiment on a FastGRNN model \cite{Kusupati2018FastGRNNAF} on the HAR-2 Human Activity Recognition dataset \cite{HAR}. HAR-2 dataset is a binarized version of the 6-class Human Activity Recognition dataset. From the full-rank model with $r_W = 9$ and $r_U = 80$ as suggested on the STR paper \cite{pmlr-v119-kusupati20a}, we apply Global MP on the matrices $W_1$ and $W_2$. To do this, we find the weight mask by ranking the columns of $W_1$ and $W_2$ based on their absolute sum, then we prune the $9 - r_W^{new}$ lowest columns and $80 - r_U^{new}$ lowest columns from $W_1$ and $W_2$ respectively. In the end, we fine-tune this pruned model by retraining it with FastGRNN's trainer and applying the weight mask at every epoch. We test Global MP under different $r_w$-$r_v$ configurations. We find that Global MP surpasses the other baselines on all the configurations (Table \ref{table:fastgrnn_har2}) and successfully prunes the model on a very different architecture and domain.

\subsection{Mitigating layer-collapse}
\label{high_sparsity}
Layer-collapse is an issue that many pruning algorithms run into \cite{NEURIPS2020_46a4378f, Lee2020A, hayou2021robust} and occurs when an entire layer is pruned by the pruning algorithm, rendering the network untrainable. We investigate this phenomena and find that the performance of a pruning algorithm can be substantially affected by the architecture of the neural network being pruned, especially in the high sparsity domain. We conduct experiments on MobileNet-V2 and WRN-22-8 models over the CIFAR-10 dataset. We report results averaged over multiple runs where each run uses a different pre-trained model to provide more robustness. We first prune a WRN-22-8 model to 99.9\% sparsity. We find that at 99.9\% sparsity, the WRN is still able to get decent accuracy (Table \ref{table:highsparsity_wrn}). We then prune a MobileNet-V2 model to 98\% sparsity. For MobileNet, however, accuracy drops to 10\% using only Global MP, and the model is not able to learn (Table \ref{table:highsparsity_mnet}).

\begin{figure}[h!]
    \centering
    \includegraphics[trim=3cm 2.7cm 12cm 2cm,clip,width=0.48\textwidth]{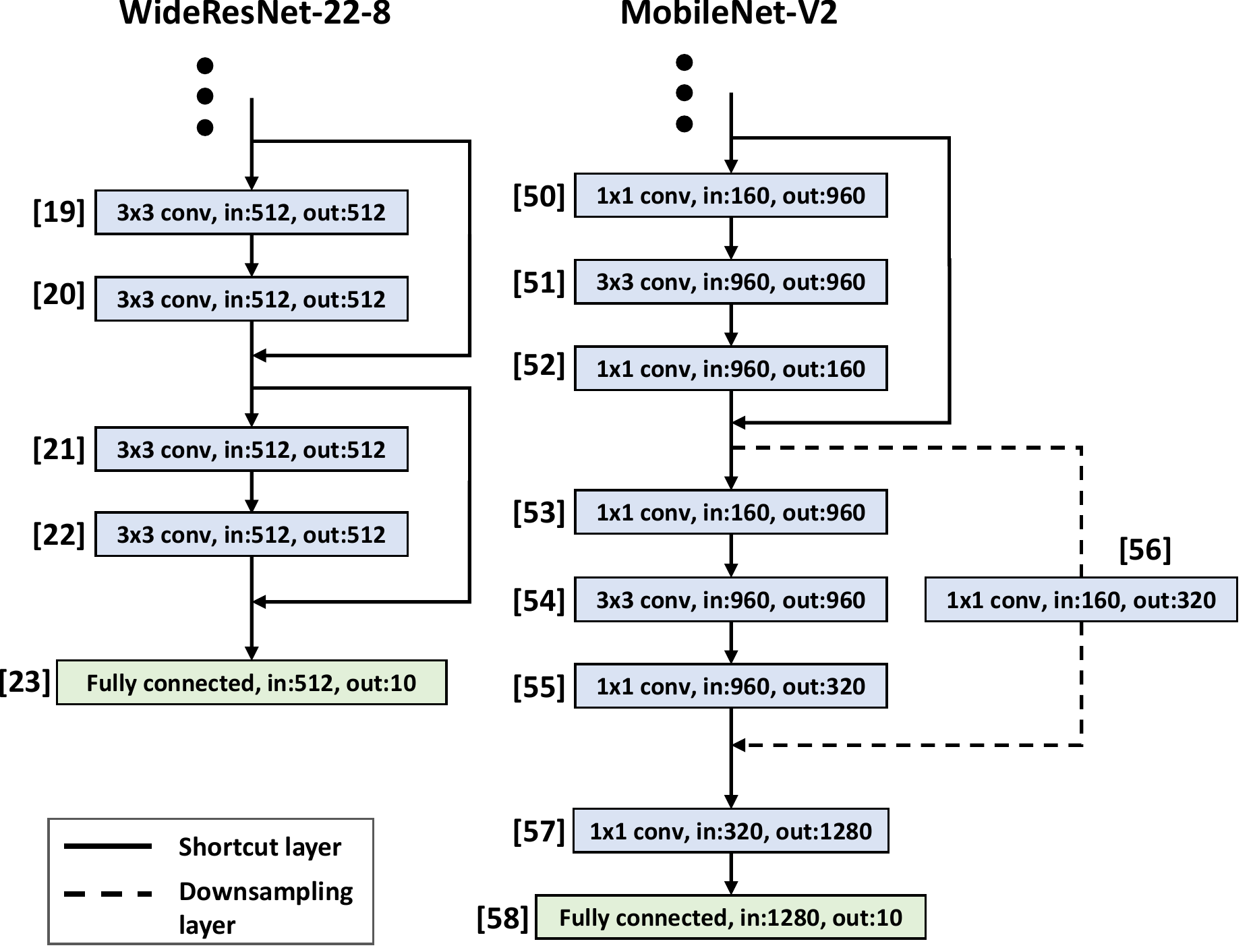}
    \caption{Difference in architectures between WRN and MobileNet. WRN does not have prunable residual connections in the last layers (dotted lines) while MobileNet does. This leads to different pruning behaviors on the two architectures.}
    \vspace*{-0.3cm}
    \label{fig:highsparsity_archs}
\end{figure}

The reason for this wide discrepancy in learning behavior lies in the shortcut connections \cite{DBLP:journals/corr/HeZRS15}. Both WRN-22-8 and MobileNet-V2 use shortcut connections, however, their placement is different. Referring to Fig. \ref{fig:highsparsity_archs}, WRN uses identity shortcut connections from Layer 20 to Layer 23. This type of shortcut connections are simple identity mappings and do not require any extra parameters, and hence, they do not count towards the weights. However, MobileNet-V2 uses a convolutional shortcut mapping from Layer 52 to Layer 57. The weights in this mapping are counted towards to the model's weights, and thus, they are prunable. Global MP completely prunes the two preceding layers before the last layer. However, because WRN uses identity mappings, it is still able to relay information to the last layer, and the model is still able to learn, whereas MobileNet-V2 faces catastrophic accuracy drop due to layer-collapse. Pruning algorithms can be susceptible to such catastrophic layer-collapse issues especially in the high sparsity domain. The MT rule can help overcome this issue. Retaining a small MT of 0.02\% is sufficient for MobileNet-V2 to avoid layer-collapse and learn successfully. Hence, retaining a small amount of weights can help in the learning dynamics of models in high sparsity settings.

\begin{table}[t!]
\centering
\small
\begin{tabular}{p{2.6cm}p{1.45cm}p{0.5cm}p{0.5cm}}
\toprule
\multirow{1}{*}{Method} & Top-1 Acc & $r_W$ & $r_U$\\
\midrule
FastGRNN & 96.10\% & 9\ & 80\\
\midrule
Vanilla Training & 94.06\% & 9\ & 8\\
STR & 95.76\% & 9\ & 8\\
\textbf{Global MP} & \textbf{95.89}\% & 9 & 8\\
\midrule
Vanilla Training & 93.15\% & 9\ & 7\\
STR & 95.62\% & 9\ & 7\\
\textbf{Global MP} & \textbf{95.72}\% & 9 & 7\\
\midrule
Vanilla Training & 94.88\% & 8\ & 7\\
STR & 95.59\% & 8\ & 7\\
\textbf{Global MP} & \textbf{95.62}\% & 8 & 7\\
\bottomrule
\end{tabular}
\captionof{table}{Results on FastGRNN on HAR-2 dataset. The bold font denotes the algorithm with the best performance. Global MP outperforms other pruning algorithms.}
\label{table:fastgrnn_har2}
\end{table}

\begin{table}[t!]
\small
\begin{tabular}{p{1.5cm}p{0.8cm}p{2.2cm}p{2.2cm}}
\toprule
\multirow{2}{*}{Method} & \multicolumn{2}{c}{WRN-22-8 on CIFAR-10} \\ \cmidrule(lr){2-4}
& Sparsity & Starting Acc. & Pruned Acc.\\
\midrule
\textbf{Global MP} & 99.9\% & $94.07\% \pm 0.05\%$ & $\textbf{67.68\%} \pm \textbf{0.78\%}$ \\
\bottomrule
\end{tabular}
\captionof{table}{Performance of Global MP on WideResNet-22-8 in the high sparsity regime at 99.9\% sparsity.}
\label{table:highsparsity_wrn}
\end{table}

\begin{table}[t!]
\small
\centering
\begin{tabular}{m{1.5cm}m{0.7cm}m{2.2cm}m{2.2cm}}
\toprule
\multirow{2}{*}{Method} & \multicolumn{2}{c}{MobileNet-V2 on CIFAR-10} \\ \cmidrule(lr){2-4}
& Sparsity & Starting Acc. & Pruned Acc.\\
\midrule
Global MP & 98.0\% & $94.15\%\pm0.23\%$ & $10\%$ \textit{(Unable to learn)} \\
\textbf{Global MP with MT} & 98.0\% & $94.15\% \pm 0.23\%$ & $\textbf{82.97\%} \pm \textbf{0.57\%}$\\
\bottomrule
\end{tabular}
\captionof{table}{Adding MT enables MobileNet-V2 to learn in the high sparsity regime.}
\label{table:highsparsity_mnet}
\end{table}

%% file: 5_Discussion.tex
\section{Discussion, Limitations and Future Work}
Our observations indicate that Global MP works very well and achieves superior performance on all the datasets and architectures tested. It can work as a one-shot pruning algorithm or as a gradual pruning algorithm. It also surpasses SOTA algorithms on ResNet-50 over ImageNet on the sparsity-accuracy trade-off and sets the new SOTA results across many sparsity levels. For FLOPs-accuracy trade-off, it comes in second after STR, surpassing many SOTA techniques. At the same time, Global MP has very low algorithmic complexity and arguably is one of the simplest pruning algorithms. It is simpler than many other pruning algorithms like custom loss based regularization, RL-based procedures, heuristics-based layerwise pruning ratios, etc. It just ranks weights based on their magnitude and removes the smallest ones. This raises a key question on whether complexity is really required for pruning and according to our results it seems that complexity in itself does not guarantee good performance. Practitioners developing new pruning algorithms should thus look carefully whether complexity is adding value to their algorithm or not and surpassing baselines like Global MP.

A limitation of Global MP is that the theoretical foundations for it have not been well-established yet. Following our empirical work in this manuscript, we plan to conduct a theoretical analysis for better comprehending the dynamics of Global MP in the future. It would include finding analytical links between the magnitude of weights and their importance in a network, or even analytical relations of them to the resultant accuracy of the model. Another area for future work is jointly optimizing both weights and FLOPs during the pruning process. Currently, Global MP is used to reach a certain parameter sparsity, and FLOPs reduction comes as a by-product. In the future, FLOPs can also be added to an optimization function to jointly sparsify both parameters and FLOPs.

%% file: 6_Conclusion.tex
\section{Conclusions}
\label{conclusion}

In this work, we raised the question of whether utilizing complex and computationally demanding algorithms are really required to achieve superior DNN pruning results. This stemmed from the hike in the number of new pruning algorithms proposed in the recent years, each with a marginal performance increment, but increasingly complicated pruning procedures. This makes it hard for a practitioner to select the correct algorithm and the best set of algorithm-specific hyper-parameters for their application. We benchmarked these algorithms against a naive baseline, namely, Global MP, which does not incorporate any complex procedure or any hard-to-tune hyper-parameter. Despite its simplicity, we found that Global MP outperforms many SOTA pruning algorithms over multiple datasets, such as CIFAR-10, ImageNet, and HAR-2; with different network architectures, such as ResNet-50 and MobileNet-V1; and at various sparsity levels from 50\% up to 99.9\%. We also presented a few variants of Global MP, i.e., one-shot and gradual, together with a new, complementary technique, MT. While our results serves as an empirical proof that a naive pruning algorithm like Global MP can achieve SOTA results, it remains as a promising future research direction to shed light into theoretical aspects of how such performance is possible with Global MP. Another future direction includes extending the capabilities of Global MP, such as jointly optimizing both FLOPs and the number of weights.

\section{Acknowledgement}

This research is supported by the Agency for Science, Technology and Research (A*STAR) under its AME Programmatic Funds (Project No: A1892b0026 and A19E3b0099). Any opinions, findings and conclusions or recommendations expressed in this material are those of the author(s) and do not reflect the views of the A*STAR.

%% file: 7_Appendix.tex
\newpage
\clearpage

\appendix
\section{Supplementary Materials}
\label{appendix}
\setcounter{figure}{2} 
\setcounter{table}{6} 

\subsection{Pseudocode}
\label{pseudocode}

\begin{algorithm}[t!]
\caption{Global MP}
\label{algo1}
\begin{algorithmic}
\STATE{\textbf{Input:} $DNN_{init}$, pre-trained or untrained DNN}
\STATE{\hspace{1.1cm}$\kappa_{target}$, target sparsity}
\STATE{\hspace{1.1cm}$\sigma$, minimum threshold (MT)}
\STATE{\hspace{1.1cm}$e_{total}$, total epochs}
\STATE{\hspace{1.1cm}$isGradual$, gradual or one-shot}
\STATE{\hspace{1.1cm}$isMT$, MT is applied or not}
\STATE{\textbf{Output:} $DNN_{final}$, pruned and trained DNN}
\vspace{0.35cm}
\STATE{$e = 0$}
\STATE{$DNN(w_e)$ = $DNN_{init}$}
\WHILE{$e < e_{total}$}
\IF{$\kappa_{DNN(w_e)} < \kappa_{target}$}
\STATE{$t_e$ $\leftarrow$ $CalcThreshold(e, \kappa_{target}, isGradual)$}
\IF{$isMT$}
\STATE{$DNN(w_{e'}) \hspace{0.1cm} \leftarrow Mask(DNN(w_e), t_e)$}
\STATE{$W \hspace{1.43cm} \leftarrow MTcheck(DNN(w_{e'}), \sigma)$}
\STATE{$DNN(w_{e^+}) \leftarrow MTprune(DNN(w_{e'}), W)$}
\ELSE
\STATE{$DNN(w_{e^+}) \leftarrow Prune(DNN(w_e), t_e)$}
\ENDIF
\ELSE
\STATE{$DNN(w_{e^+}) = DNN(w_e)$}
\ENDIF
\STATE{$DNN(w_{e^{++}}) \leftarrow BackProp(DNN(w_{e^+}))$}
\STATE{$DNN(w_e) = DNN(w_{e^{++}})$}
\STATE{$e = e+1$}
\ENDWHILE
\STATE{$DNN_{final}$ = $DNN(w_e)$}
\end{algorithmic}
\end{algorithm}

The pruning pipeline for Global MP (Algorithm \ref{algo1} and Table \ref{table:func_list}) starts by first taking a pre-trained model for the case of one-shot pruning or untrained model for the case of gradual pruning. Next, the sparsity of the model is checked and if the sparsity is lower than the target sparsity, then the model is pruned using either vanilla Global MP or Global MP with MT, as per the choice of the user. Once, the model is pruned, it is trained for the case of gradual pruning or fine-tuned for the case of one-shot pruning. The above procedure repeats until the final epoch is reached. For the case of one-shot pruning, the later epochs are just used for doing fine-tuning as the pruning happens in one-go in the first epoch itself. This finishes the procedure and the final result is a pruned and trained (or fine-tuned) model. 

\begin{table}[t!]
\caption{Function explanations for Algorithm 1.}
\label{table:func_list}
\centering
\begin{tabular}{p{2.5cm}p{5cm}}
\hline\noalign{\smallskip}
\textbf{Function} & \textbf{Explanation} \\ 
\noalign{\smallskip}\hline\noalign{\smallskip}
\multirow{4}{*}{$CalcThreshold()$:} & Calculates the magnitude threshold below which \\
& the weights are to be pruned. Assigns all pruning \\
& budget in one epoch or distributes it to epochs \\
& based on $isGradual$. \\
\noalign{\smallskip}
\multirow{2}{*}{$Mask()$:} & Identifies the weights to be pruned without \\
& actually pruning them. \\
\noalign{\smallskip}
\multirow{3}{*}{$MTcheck()$:} & Checks the layers that violate the MT condition \\
& and returns a new mask by distributing \\
& the pruning budget among other layers. \\
\noalign{\smallskip}
\multirow{2}{*}{$MTprune()$:} & Prunes based on the mask returned by \\
& $MTcheck()$. \\
\noalign{\smallskip}
$Prune()$: & Prunes based on a threshold. \\
\noalign{\smallskip}
$BackProp()$: & Conducts a single back-propagation for training. \\
\hline
\vspace{-1cm}
\end{tabular}
\end{table}

\subsection{MT Ablation}
\label{ablation}

\begin{figure}[h!]
    \centering
    \includegraphics[trim={0.4cm 0.2cm 0.4cm 0.2cm}, clip, width=0.44\textwidth]{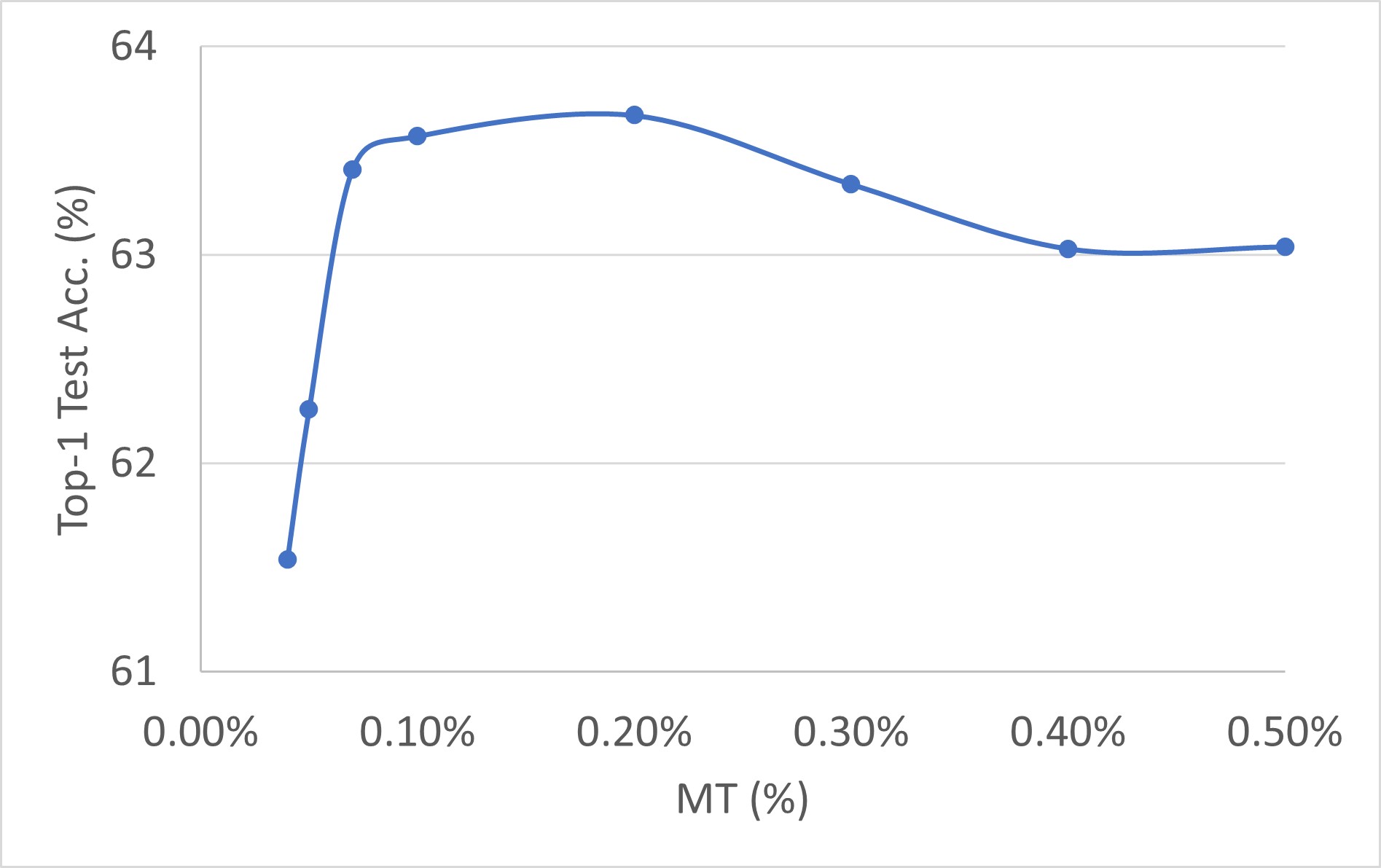}
    \caption{Ablation study on how changing the minimum threshold (MT) affects accuracy. Just like tuning any hyper-parameter, the accuracy increases when MT is initially increased until it hits a maximum value, after which the accuracy decreases on increasing MT further. Hence, MT is easy to search and can be set in the same way as searching for any hyper-parameter. Using MobileNet-V1 on ImageNet.}
    \vspace*{-0.3cm}
    \label{fig:mt_ablation}
\end{figure}

We provide an ablation study for the minimum threshold (MT) (Fig. \ref{fig:mt_ablation}). We find that MT behaves like a typical hyper-parameter. On increasing the MT value initially, the accuracy increases, until it reaches a maximum value. Thereafter, increasing the MT value leads to a decrease in accuracy. Thus, a suitable value can be found by doing a search over MT values. Usually values between 0.01\% to 0.3\% work well for MT regardless of models and datasets. 

\subsection{Layer-wise results}
\label{layerwise}

\begin{figure*}[h!]
    \centering
    \includegraphics[trim=1.5cm 11cm 1.5cm 11cm,clip,width=\textwidth]{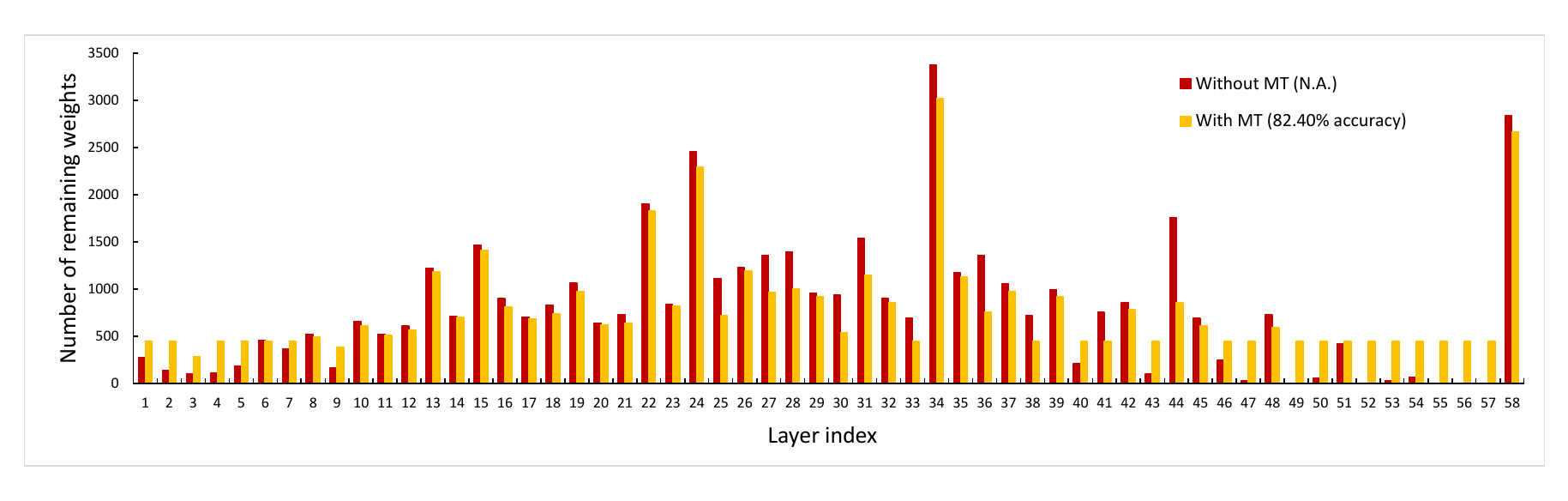}
    \vspace*{-0.9cm}
    \caption{For MobileNet-V2 at 98\% sparsity, MT helps retain some weights in the heavily pruned layers (Layers 55, 56, and 57) and allows the model to learn successfully.}
    \vspace*{-0.3cm}
    \label{fig:highsparsity_mnet}
\end{figure*}

\begin{figure*}[h!]
    \centering
    \includegraphics[trim=1.5cm 11cm 1.5cm 11cm,clip,width=\textwidth]{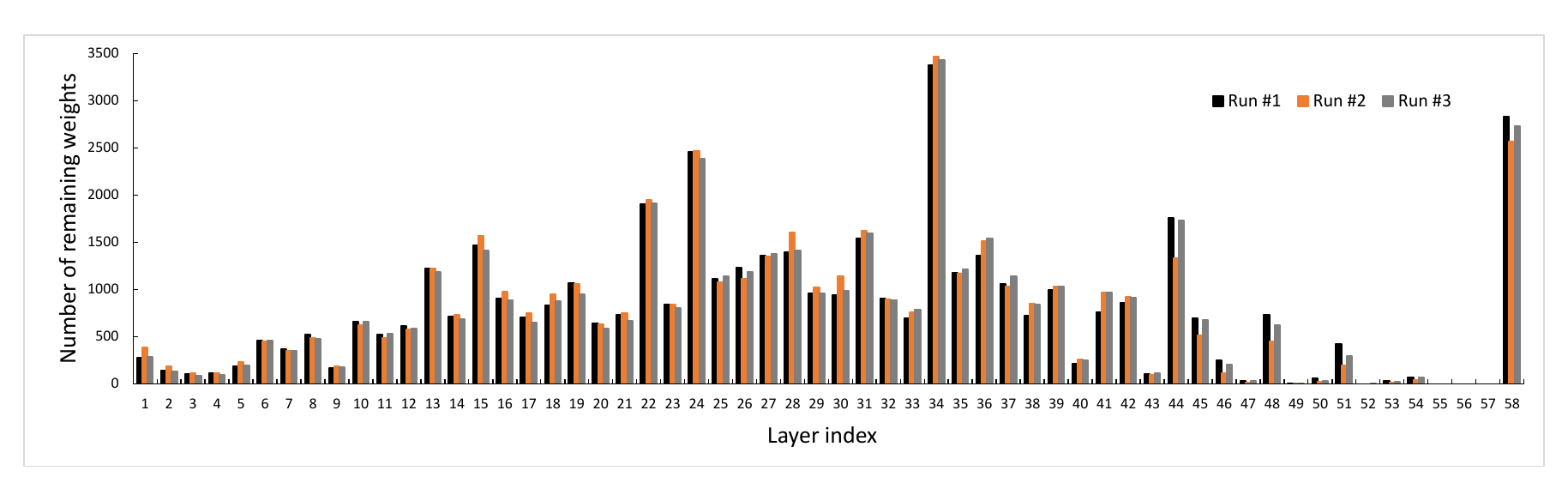}
    \vspace*{-0.9cm}
    \caption{Layer-wise pruning results produced by Global MP on MobileNet-V2 model on CIFAR-10. Pruning is conducted on three different pre-trained models and the pruning results across the three runs are very stable.}
    \vspace*{-0.3cm}
    \label{fig:highsparsity_mnet_gp}
\end{figure*}

Pruning algorithms are susceptible to catastrophic layer-collapse issues especially in the high sparsity domain. The MT rule can help overcome this. Applying MT helps MobileNet-V2 model to avoid layer-collapse and learn successfully. MT achieves this by preserving a critical mass of weights in the later layers of MobileNet-V2 (Fig. \ref{fig:highsparsity_mnet}). Hence, retaining a small amount of weights can help in the learning dynamics of models in high sparsity settings. We also plot the layer-wise snapshot for Global MP and find that it is very stable and produces consistent pruning results across multiple runs and pre-trained models (Fig. \ref{fig:highsparsity_mnet_gp}).

\subsection{Sparsifying FastGRNN on HAR-2 dataset}
\label{fastgrnn}
HAR-2 dataset that is used in the FastGRNN pruning experiment is a binarized version of the 6-class Human Activity Recognition dataset. From the full-rank model with $r_W = 9$ and $r_U = 80$ as suggested on the STR paper \cite{pmlr-v119-kusupati20a}, we apply GP on the matrices $W_1$ and $W_2$. To do this, we find the weight mask by ranking the columns of $W_1$ and $W_2$ based on their absolute sum, then we prune the $9 - r_W^{new}$ lowest columns and $80 - r_U^{new}$ lowest columns from $W_1$ and $W_2$ respectively. In the end, we fine-tuned this pruned model by retraining it with FastGRNN's trainer and applying the weight mask on every epoch.

\subsection{Hyper-parameters}
\label{hyperparams}

For CIFAR-10, we use batch-size of 128, momentum 0.9, 300 epochs, weight decay between 0 to 0.001, learning rate between 0.01 to 0.1 with cosine decay and nesterov momentum for WRN-28-8. For ImageNet, we use batch-size of 256, momentum 0.875, 100 epochs for ResNet-50 and 120 for MobileNet-V1, weight decay between 0 to 3.1e-5, learning rate between 0.0256 to 0.256 with cosine decay, label smoothing 0.1 and 5 epochs of warm-up for gradual pruning. Averages reported over three runs.